\documentclass[twoside,10pt,reqno]{article}
\usepackage[english]{babel}
\usepackage[utf8]{inputenc}

\usepackage[pdftex,colorlinks=true]{hyperref} 

\usepackage{amsmath}
\usepackage{amsfonts}
\usepackage{amssymb}
\usepackage{anysize}
\usepackage{enumerate}
\usepackage{bbm} 
\usepackage{xcolor}
\usepackage[normalem]{ulem}
\usepackage{xargs}                      
\definecolor{bluee}{HTML}{4967ed}
\definecolor{malgared}{rgb}{0.8627450980392157, 0.34901960784313724, 0.34901960784313724}
\usepackage{tikz}
\usetikzlibrary{
	calc,
	matrix,
	fit,
	positioning,
	decorations,
	decorations.pathreplacing,
	decorations.pathmorphing,
	arrows.meta,
}

\DeclareRobustCommand{\SkipTocEntry}[5]{}

\usepackage{kpfonts}
\usepackage{xspace}
\usepackage{bbm}
\newcommand{\ind}{\mathbbm{1}}

\newcommand{\knm}{\widehat{K}_{nM}}
\newcommand{\kmn}{\widehat{K}_{nM}^\top}
\newcommand{\kmm}{\widehat{K}_{MM}}

\def\argmin{\operatornamewithlimits{arg\,min}}

\DeclareMathOperator{\E}{\mathbb{E}}
\renewcommand{\P}{\mathbb{P}}

\newcommand{\wh}{\widehat} 
\newcommand{\wt}{\widetilde}

\newcommand{\eni}{\begin{equation}}
\newcommand{\enf}{\end{equation}}
\newcommand{\be}{\begin{equation}}
\newcommand{\ee}{\end{equation}}
\newcommand{\br}{\begin{rem}}
\newcommand{\er}{\end{rem}}
\newcommand{\bt}{\begin{thm}}
\newcommand{\et}{\end{thm}}
\newcommand{\bp}{\begin{prop}}
\newcommand{\ep}{\end{prop}}
\newcommand{\bex}{\begin{ex}}
\newcommand{\eex}{\end{ex}}
\newcommand{\bd}{\begin{defi}}
\newcommand{\ed}{\end{defi}}

\newcommand{\bn}{\begin{enumerate}}
\newcommand{\en}{\end{enumerate}}
\newcommand{\bi}{\begin{itemize}}
\newcommand{\ei}{\end{itemize}}
\newcommand{\beas}{\begin{eqnarray*}}
\newcommand{\eeas}{\end{eqnarray*}}
\newcommand{\bea}{\begin{eqnarray}}
\newcommand{\eea}{\end{eqnarray}}

\newtheorem{thm}{Theorem}
\newtheorem{defi}{Definition}

\newtheorem{prop}{Proposition}
\newtheorem{rem}{Remark}
\newtheorem{ex}{Example}

\providecommand{\nor}[1]{\left\lVert {#1} \right\rVert}

\providecommand{\scal}[2]{\left\langle{#1},{#2}\right\rangle}

\newcommand{\R}{\mathbb R}

\newcommand{\N}{\mathbb N}

\newcommand{\XX}{\mathcal X}

\newcommand{\hh}{\mathcal H}
\newcommand{\cL}{\mathcal L}
\newcommand{\cB}{\mathcal B}

\newcommand{\la}{\lambda}

\newcommand{\X}{\XX}
\newcommand{\Y}{\mathcal Y}
\newcommand{\Z}{\mathcal Z}

\newcommand{\G}{\mathcal G}

\newcommand{\pz}[1]{\textcolor{violet}{#1}}

\newcommandx{\unsure}[2][1=]{\todo[linecolor=red,backgroundcolor=red!25,bordercolor=red,#1]{#2}}

\title{Learning functions, operators and dynamical systems with kernels}
\author{Lorenzo Rosasco}

\date{}

\begin{document}
	\maketitle
\begin{abstract}
This expository article presents the approach to statistical machine learning based on reproducing kernel Hilbert spaces. The basic framework is introduced for scalar-valued learning and then extended to operator learning. Finally, learning dynamical systems is formulated as a suitable operator learning problem, leveraging Koopman operator theory.  The manuscript collects  supporting material for the corresponding course taught at the CIME school "Machine Learning: From Data to Mathematical Understanding" in Cetraro.

\end{abstract}

\section{Learning from data}
The most fundamental problem in machine learning is estimating a function $\wh f$ from data, i.e., pairs of inputs and outputs $(x_1, y_1), \dots, (x_n, y_n)$. The key intuition is that the function $\wh f$ should give an estimate of the output for any {\em new} input, ideally satisfying $\wh f (x_{\text{new}})\approx y_{\text{new}}$. The function to be learned can be interpreted as a task and the data as experience from which solving the task can be learned. The term ``learning" arises from this perspective. This idea is the so-called learning-from-examples paradigm. We illustrate the generality of this approach with three different examples.

\begin{figure}[h]
	\centering
	\includegraphics[width=0.5\textwidth]{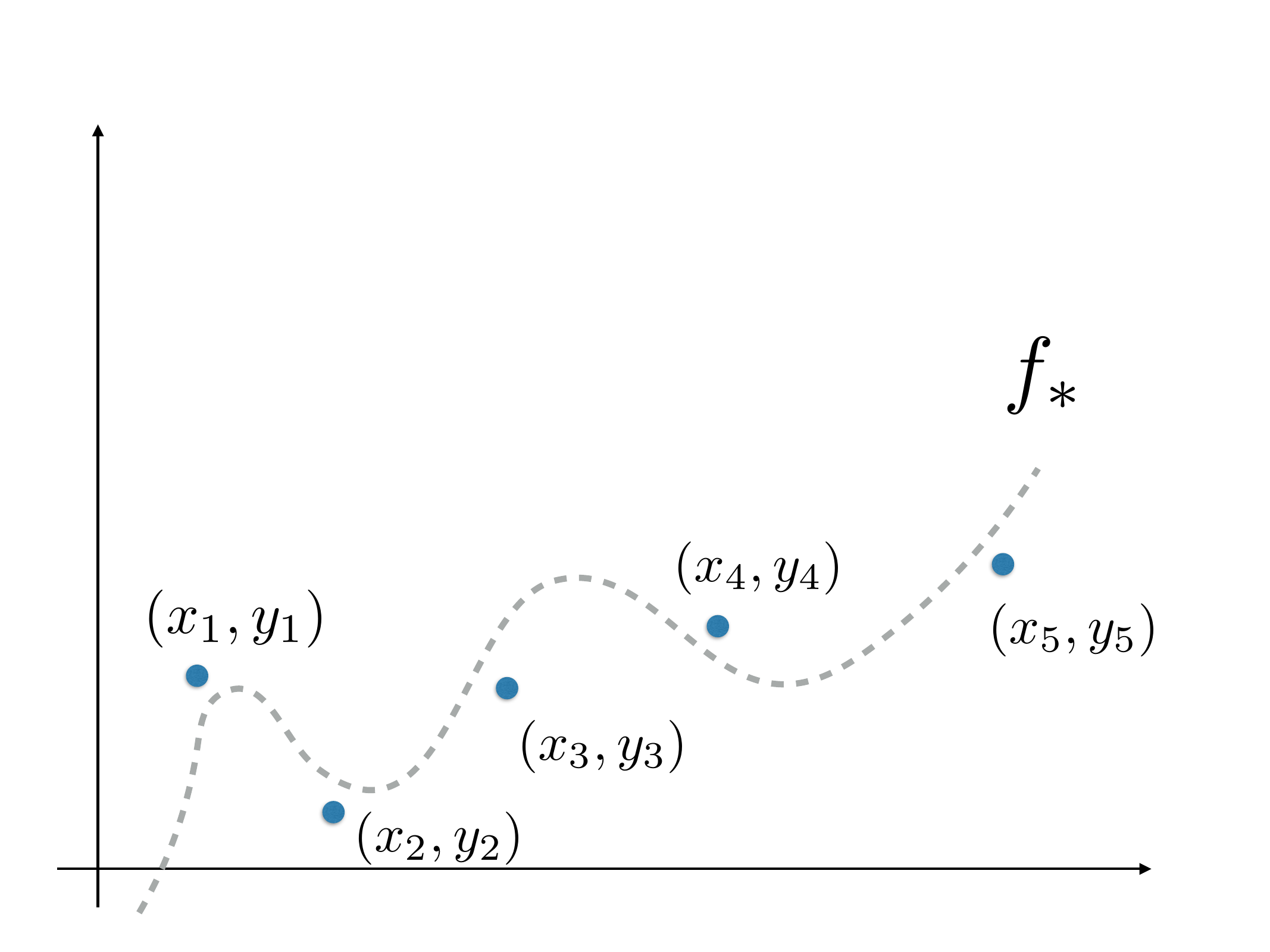}
	\caption{Supervised learning is the problem of finding an estimate $\widehat{f}$ of  a function $f_*$ given  data $(x_i, y_i)_{i=1}^n$.}
	\label{fig:data}
\end{figure}

\bex[Regression]
Regression is a supervised learning problem where the outputs are scalar-valued. More precisely, consider the following data model, 
$$
y_i=f_*(x_i)+\epsilon_i\qquad i=1, \dots, n.
$$
Here, for any $i=1, \dots, n$, $x_i\in \R^d$, $d\ge 1$, $y_i\in \R$, $f_*:\R^d\to\R$ is a unknown function to be learned, and, for any $i=1, \dots, n$, $\epsilon_i \sim \mathcal{N}(0,\sigma^2)$. Both the function $f_*$ and the noise are fixed but unknown. Given the data $(x_i, y_i)_{i=1}^n$, the goal is to compute an estimate $\wh f$ of $f_*$. This model is illustrated in Figure~\ref{fig:data}. As an example, we can think of $x_i$ as a set of features describing a house (e.g., size, number of rooms, location) and $y_i$ as its market price. The task is to learn to predict house prices from given features.
\eex

\bex[Operator learning]\label{oplearn}
Operator learning is a supervised learning problem where inputs and outputs belong to Hilbert (or Banach) spaces. More precisely, consider the following model,
$$
y_i=F_*(x_i)+\epsilon_i\qquad i=1, \dots, n.
$$
Here, for any $i=1, \dots, n$, $x_i \in \mathcal{X}$ and $y_i \in \mathcal{Y}$, where $\mathcal{X}$ and $\mathcal{Y}$ are Hilbert spaces. The function to be learned is a fixed but unknown linear operator $F_*: \mathcal{X} \to \mathcal{Y}$, and for any $i=1, \dots, n$, the noise term $\epsilon_i$ is a zero-mean random variable. Given the data $(x_i, y_i)_{i=1}^n$, the goal is to compute an estimate $\wh F$ of $F_*$. As an example, we can think of $y_i$ as images and $x_i$ as their blurred, noisy versions. The task is to learn to reconstruct sharp images from their degraded versions.
\eex

\bex[Dynamical systems learning]\label{dslearn}
Dynamical systems learning is a supervised learning setting where the goal is to estimate a state transition function from observed trajectories. More precisely, consider the following model, for some initial condition $s_0$,
$$
s_{t+1}= f_*(s_t, \omega_t) \qquad t = 0, \dots, T-1.
$$
Here, for any $t = 0, \dots, T-1$, $s_t \in \mathcal{S}=\R^d$ is the system state, $\omega_t$ is a random variable modeling stochasticity, and $f_*: \mathcal{S} \times \Omega \to \mathcal{S}$ is a fixed but unknown function describing the state evolution. Given a trajectory $(s_0, s_1, \dots, s_{T-1})$, the goal is to compute an estimate $\wh f$ of $f_*$. As an example, we can think of $s_t$ as the position and velocity of a robot and $\omega_t$ as external disturbances, such as sensor noise or environmental factors. The task is to learn to predict the robot's next state based on past observations.
\eex

The above examples show that  Figure \eqref{fig:data} provides a simple but potentially misleading illustration. 
The figure depicts a low-dimensional setting, real-world data can have tens, or even tens of thousands, of dimensions, and datasets can contain millions or even billions of samples.

Beyond these examples and observations, ensuring that the learning problem is well-posed requires assumptions about the data. Available data must be informative about new data. As suggested in the previous examples, one possible assumption is the existence of a data-generating process. This is the perspective taken in the framework of statistical learning theory, which we describe next.

\section{Statistical Learning}

In this section, we introduce the problem of supervised machine learning within the framework of statistical learning theory. For the time being we focus on scalar valued learning. 

\paragraph{The statistical learning problem.}
Let $(\mathcal X, \mathcal A)$ be a measure space with $\sigma$-algebra $\mathcal A$, 
and let $\mathcal Y\subseteq \R$ with the corresponding Borel $\sigma$-algebra $\mathcal B$.  
Let $(X,Y)$ be a pair of random variables taking values in $(\mathcal X \times \mathcal Y)$ with law $P$.  
Let $\ell: \mathcal Y\times \mathcal Y \to [0,\infty)$ be a given measurable function.  

Consider ${\mathcal M} ({\mathcal X}, \R) \subset \R^{\mathcal X}$, the space of all measurable functions from $\mathcal X$ to $\R$.  
Define $L:{\mathcal M} ({\mathcal X}, \R)\to [0,\infty)$ as
$$
L(f) = \E[\ell(Y,f(X))], \quad \forall f\in {\mathcal M} ({\mathcal X}, \R).
$$
The problem of learning is to solve 
\be\label{scal_prob}
\min_{f\in \mathcal  M (\mathcal  X, \R)} L(f),
\ee
where $P$ is only known through a sample $(x_i,y_i)_{i=1}^n$ of $n$ independent and identically distributed copies of $(X,Y)$.

This framework is referred to as statistical supervised machine learning.  
Given data $(x_i,y_i)_{i=1}^n$ drawn from a distribution, the goal is to find a function with small error on future data sampled from the same distribution.  
The error is measured pointwise by the loss $\ell$, while the expected risk $L$ represents the error over {\em all} possible future data.  

In general, finding a perfect solution  is infeasible since the data distribution is unknown;  but  one can  find empirical solutions that become more accurate as more data become available.  
The problem is supervised since each input $x_i$ has a corresponding output $y_i$.  

Next, we discuss the interpretation of the quantities introduced above and provide examples.  
Later, we describe how empirical solutions can be found.
\subsection{Data Space and Distribution}
We next discuss the interpretation and examples of the probability space $(\mathcal X \times \mathcal Y, {\mathcal A}\otimes {\mathcal B}, P)$ where the data reside.

\paragraph{Input and output spaces.}  
The space $\mathcal X$ is called the input space. An example is ${\mathcal X} \subseteq \R^d$ with the corresponding Borel $\sigma$-algebra. More generally, $\mathcal X$ can be a metric space equipped with the corresponding Borel $\sigma$-algebra. For example, $\mathcal X$ could be the space of binary strings $\{0,1\}^d$ with the Hamming distance  
$$
d_{\text{Ham}}(x,\bar x)=
\sum_{j=1}^d \ind_{x^j\neq \bar x^j},
$$ 
or the simplex  
$
\{x \in \R^d~|~\sum_{j=1}^d x^j=1~\text{and}~x^j\ge 0~\text{for}~j=0, \dots, d\}
$  
with the Hellinger distance  
$$
d_{\text{Hel}}(x,\bar x)=
\frac 1 2 \sqrt{\sum_{i=1}^d (\sqrt{x^j}-\sqrt{\bar x^j})^2},
$$  
or another metric on probability distributions.

The space $\mathcal Y \subseteq \R$ is called the output space. A special case is when $\mathcal Y \subset \R$, and in particular, if $\mathcal Y$ takes only two values, such as $\{-1,1\}$, the learning problem is referred to as binary classification.  

\paragraph{Data distribution.}  
The distribution $P$ is a probability measure on the space $(\mathcal X \times \mathcal Y, \mathcal A \otimes \mathcal B)$. The marginal measure $P_X$ on $(\mathcal X, \mathcal A)$ is defined for all $A \in \mathcal A$ as  
$$
P_X(A) = \int_{A\times {\mathcal{Y}}} d P(x,y).
$$  
For all $x\in \mathcal X$, there exists a Borel probability measure $P(\cdot~|x)$ on $(\mathcal Y, \mathcal B)$, known as the conditional measure at $x$, such that for all $B\in {\mathcal B}$, the function $x\mapsto P(B~|x)$ is measurable, and for all $A\in {\mathcal A}$ and $B\in {\mathcal B}$,  
$$
P(A\times B) = \int P(B~|~x) dP_X(x).
$$  
In binary classification, the conditional distribution reduces to the point mass function $\{P(1~|~x), P(-1~|~x)\}$. The general proof of decomposition in marginal and conditional probability measures is technical, see bibliography. It becomes elementary for discrete random variables and for continuous random variables with a well-defined probability density function. In the latter case, the existence of a well-defined conditional distribution follows from Fubini's theorem.  

\paragraph{Training set.}  
The set of pairs $(x_i,y_i)_{i=1}^n$ is referred to as the training set. The acronym i.i.d. (independent and identically distributed) is commonly used to describe such samples.  
While the i.i.d. assumption is strong, it is the standard one used to derive fundamental results.

\subsection{Loss Function, Expected Risk, and Target Function}
Since we consider deterministic functions within a probabilistic setting, we must account for possible errors. This motivates the introduction of the loss function and the corresponding expected loss (risk). 

\paragraph{Loss function.}  
The function $\ell$ is called the loss function. It can be viewed as a pointwise error measure for any pair of outputs. In particular, $\ell(y, f(x))$ represents the error incurred when predicting $f(x)$ instead of $y$. It is often assumed to be continuous and convex in the second argument. The primary loss function we consider is the least squares loss
$$
\ell(y,y')= (y-y')^2
$$
for all $y,y'\in \Y$.  

More generally, in regression, loss functions take the form $\ell(y,y')= V(y-y')$ for all $y,y'\in \R$, where $V:\R\to [0,\infty)$. An example, besides the square loss, is the absolute value loss $V(a)= |a|$, $a\in \R$. In binary classification, loss functions are of the form $\ell(y,y')= V(yy')$ for all $y,y'\in \R$, with $V:\R\to [0,\infty)$. Examples include the misclassification loss $V(a)= \ind_{a<0}$, the square loss $V(a)=(1-a)^2$, the logistic loss $V(a)=\log(1+e^{-a})$, the exponential loss $V(a)=e^{-a}$, and the hinge loss $V(a)= |1-a|_+=\max\{0, 1- a\}$, for $a\in \R$.

\paragraph{Expected risk and target function.}  
For all $f\in\mathcal{M}$, the functional $L: \mathcal{M}(\X, \R) \to [0, \infty)$ defined as
\begin{equation}\label{exprsk}
	L(f) = \E[\ell(Y, f(X))]= \int \ell(y, f(x)) dP(x, y)
\end{equation}
is called the expected loss or expected risk. Assume that there exists $f_P\in {\mathcal  M} ({\mathcal  X}, \R)$ such that 
$$
L(f_P)= \min_{f\in {\mathcal  M} ({\mathcal  X}, \R)}L(f),
$$
then $f_P$ is called a target function.  

The expected risk can be interpreted as the error over all possible input-output pairs as measured by the loss function, where errors are weighted more for pairs that are more likely to be sampled. The target function can be seen as an ideal solution achieving the best possible error on all future data.  

In practice, only empirical approximations of the target function can be computed. However, as describe next, analytic expressions can be derived in terms of the underlying probability distribution, providing insights into the problem.  
\paragraph{Inner risk and target function.}  
For almost all $x\in \mathcal  X$, let  $L_x: \R\to  [0,\infty)$ be defined as $L_x(a)=\int \ell(y,a)dP(y~|~x)$ for all $a\in \R$. The function $L_x$ is referred to as the inner risk at $x$.  
Note that, for all $f\in {\mathcal  M} ({\mathcal  X}, \R)$,  
\begin{equation*}
	{L(f) = \int \ell(y, f(x)) dP(x, y) = \int \Bigg( \int \ell(y, f(x))dP(y|x) \Bigg) dP_X(x) = \int L_x(f(x))dP_X(x).}
\end{equation*}
The inner risk is useful for characterizing the minimizers of the risk analytically. Indeed, assume that, for almost all $x\in \mathcal  X$, there exists  $a_x\in  \R$ such that  
\be\label{inner_target}
L_x(a_x)= \min_{a\in \R}L_x(a).
\ee
Then, defining  
\be\label{inner_target2}
f_P(x)= a_x,
\ee 
for almost all $x\in {\mathcal  X}$, it is possible to show that $f_P\in {\mathcal  M} ({\mathcal  X}, \R)$ and that $f_P$ is a target function.  
It is an exercise to prove that $f_P$ as in Equation~\eqref{inner_target2} is indeed a target function. Proving that it is measurable relies on technical facts in measure theory, see bibliography.  

We illustrate the usefulness of the above results by deriving the target function for the square loss.

\bex[Regression function]  
The target function for the square loss is called the regression function and is given, for almost all $x\in \mathcal  X$, by  
$$
f_P(x)=\int ydP(y~|~x),
$$
that is, the conditional mean of $y$ given $x$.  
To see this, recalling~\eqref{inner_target} and~\eqref{inner_target2}, we verify that  
\begin{equation*}
	 f_P(x) = \argmin_{a\in\R} \int (y - a)^2 dP(y|x) = \int y dP(y~|~x).
\end{equation*}
\eex
Analogous computations can be considered for other loss functions and are left to the interested reader. Next, we move on to discuss the main approach to derive empirical solutions, namely empirical risk minimization.

\subsection{Empirical risk minimization and hypothesis spaces}
The basic intuition behind empirical risk minimization (ERM) is to take the problem in Equation~\eqref{scal_prob}, replace the expectation with an average over the available data, i.e., the empirical risk, and restrict the space of candidate functions from $\mathcal{M}(\mathcal{X}, \R)$ to a smaller space, allowing for efficient computations.

\paragraph{Empirical risk.}  
Given a training set $(x_i,y_i)_{i=1}^n$ and a loss function $\ell$, define $\wh L :  {\mathcal  M} (\X, \R)\to [0,\infty)$ by  
\be\label{emp_risk}
\wh L (f)=\frac 1 n \sum_{i=1}^n \ell(y_i, f(x_i)).
\ee 
The functional $\wh L$ is referred to as the empirical risk, also known as the training error.  

\paragraph{Hypothesis space.}  
Let $\hh\subset  {\mathcal  M} (\X, \R)$ denote a subspace of  candidate functions (hypotheses) from which a solution is selected.  
$\hh$ is referred to as the hypothesis space.  

\paragraph{Empirical risk minimization (ERM).}  
Consider the problem  
$$
\min_{f\in \hh} \wh L(f).
$$  
In general, both the computation of a solution and its properties depend on the chosen hypothesis space $\hh$.  
Next, we discuss some basic examples.

\subsubsection{Examples of hypothesis spaces}
We provide some basic examples of hypothesis spaces.  

\bex[Linear functions]  
Let $\mathcal{X} = \mathbb{R}^d$ and  
\be\label{linH}
\hh= \{f:\mathcal{X}\to \mathbb{R}~|~\exists w\in \mathbb{R}^d~\text{s.t.}~ f(x)= w^\top x, ~\forall x\in \mathcal{X}\}.
\ee
\eex  

\bex[Dictionaries of features]  
For $j=1, \dots, p$, let $\phi_j:\mathcal{X} \to \mathbb{R}$, and define  
\be\label{FeatH}
\hh= \{f:\mathcal{X}\to \mathbb{R}~|~\exists w\in \mathbb{R}^p~\text{s.t.}~ f(x) = \sum_{j=1}^p w^j \phi_j(x), ~\forall x \in \mathcal{X} \}.
\ee  
The functions $\phi_j$ are often referred to as atoms or features, and their collection is called a dictionary.  
\eex  
\bex[Neural networks]  
Let $\mathcal{X} = \mathbb{R}^d$, $\sigma:\mathbb{R}\to \mathbb{R}$, and define  
$$
\hh= \{f:\mathcal{X}\to \mathbb{R}~|~\exists \beta \in \mathbb{R}^p, w_j \in \mathbb{R}^d, j=1,\dots, p,~\text{s.t.}~ f(x)= \sum_{j=1}^p \beta^j\sigma(w_j^\top x), ~\forall x\in \mathcal{X}\}.
$$  
The function $\sigma$ is called the activation function.  
For example, $\sigma(a)=|a|_+$ is known as the rectified linear unit (ReLU), while $\sigma(a)= (1+e^{-a})^{-1}$ is called the sigmoid function.  
A linear neural network corresponds to $\sigma$ being the identity function.  The vectors $w_j$ define the so-called hidden units $\sigma(w_j^\top \cdot)$, and $p$ is the number of hidden units.  
Often, additional offsets $b_j \in \mathbb{R}, j=1,\dots, p$ are considered, in which case the functional expression becomes  
$$
f(x)= \sum_{j=1}^p \beta^j\sigma(w_j^\top x+b_j). 
$$  
Let $W:\mathbb{R}^d\to \mathbb{R}^p$ be the matrix with rows $w_1, \dots, w_p$,  
we can also write  
$$
f(x)= \beta^\top\sigma(Wx),
$$  
where, by abuse of notation, the activation function applied to a vector is understood to act component-wise.   If offsets are considered, then the matrix $W$ is replaced by a suitable affine map.  
\eex

\bex[Deep neural networks]  
Let $\mathcal{X} = \mathbb{R}^d$, $\sigma:\mathbb{R} \to \mathbb{R}$.  
Let $L \in \mathbb{N}$, and for each $\ell = 0, \dots, L-1$, let $d_\ell \in \mathbb{N}$ with $d_0 = d$. Define  
\be
\hh= \{f:\mathcal{X}\to \mathbb{R}~|~\exists \beta \in \mathbb{R}^{d_{L-1}}, W_\ell \in \mathbb{R}^{d_{\ell} \times d_{\ell-1}}, \ell=1,\dots, L-1,~\text{s.t.}~ f(x)= \beta^\top\sigma(W_{L-1} \dots W_2\sigma(W_1x)), ~\forall x\in \mathcal{X} \}.
\ee  
The above function space describes deep neural networks (DNNs), also called multi-layer perceptrons or multi-layer networks.  
Each matrix $W_\ell$ corresponds to a hidden layer, and $L$ represents the total number of layers.  
\eex

We conclude with two remarks.   First, we note that, aside from linear functions, all the above examples are spaces of nonlinear functions.   However, like linear functions, feature-based functions are linearly parameterized, whereas neural networks are nonlinearly parameterized.   Second, all the examples above describe functions defined by a finite number of parameters.   Next, we introduce reproducing kernel Hilbert spaces, which allow us to consider infinitely many parameters, and discuss the corresponding ERM problem.

%
		
\subsection{Bibliography}
	Comprehensive treatments of statistical learning theory and empirical risk minimization are available in~\cite{vapnik_slt_1998,devroye_precog_1996,gyorfi_distfree_2002}, as well as~\cite{devore_sup_2006, cucker_foundations_2002}. A standard reference on probability theory is \cite{dudley_prob_2002}.

\section{Regularized ERM in reproducing kernel Hilbert spaces} 
Kernel methods for supervised machine learning rely on ERM in a reproducing kernel Hilbert space (RKHS),   a  general class of possibly infinite-dimensional function spaces.

\subsection{Reproducing kernel Hilbert spaces}  
A reproducing kernel Hilbert space (RKHS) can be defined in several different yet equivalent ways, as discussed next.  

\paragraph{Evaluation functionals and RKHS.}  
Let $\mathcal{X}$ be a set. A reproducing kernel Hilbert space  $\hh\subset \mathbb{R}^\mathcal{X}$ is a Hilbert space with inner product $\scal{\cdot}{\cdot}_\hh$, such that for all $x \in \mathcal{X}$, the evaluation functionals $e_x: \hh\to \mathbb{R}$ defined by  
\be\label{ev1}
e_x(f)= f(x)
\ee  
are linear and continuous. Note that this implies, in particular, that  
$$
|f(x)|\lesssim \nor{f}_\hh.
$$  

We add two remarks.  

\begin{rem}[The space of continuous functions is not an RKHS]  
Let $\mathcal{X} = \mathbb{R}^d$, and consider  ${\mathcal C}(\mathcal{X})\subset \mathbb{R}^{\mathcal{X}}$, the space of continuous functions with the sup norm  
$$
\nor{f}_\infty=\sup_{x\in \mathcal{X}}|f(x)|, \quad  \forall f \in\mathcal{C}(\mathcal{X}).
$$  
Indeed, for all $f \in \mathcal{C}(\mathcal{X})$ and for all $x \in \mathcal{X}$, we have $ |f(x)|\le \nor{f}_\infty$.  
However, $\mathcal{C}(\mathcal{X})$ is not a Hilbert space, since the sup norm is not induced by an inner product.  
To verify this, we check that $\nor{f}_\infty$ violates the parallelogram law.  
\end{rem}

\begin{rem}[The space of square-integrable functions is not an RKHS]  
Consider the Hilbert space $L^2(\mathbb{R}^d) = \{f:\mathbb{R}^d \to \mathbb{R}~|~ \nor{f}^2_{L^2}= \int |f(x)|^2dx < \infty\}$,  
  with inner product  
$$
\scal{f}{f'}_{L^2} = \int f(x)f'(x)dx.
$$  
Since the norm $\nor{f}_{L^2}$ is only defined up to sets of measure zero, it does not control the value of $f$ at every $x$.  
Hence, $L^2$ is not an RKHS.  
\end{rem}  

The definition in terms of evaluation functionals highlights the generality of the notion of RKHS.  
An equivalent definition reveals additional properties.

\paragraph{RKHS and reproducing kernels.}  
A reproducing kernel Hilbert space $\hh\subset \mathbb{R}^\mathcal{X}$ is a Hilbert space with inner product $\scal{\cdot}{\cdot}_\hh$, such that there exists a function $k:\mathcal{X} \times \mathcal{X} \to \mathbb{R}$, called the reproducing kernel, satisfying the following properties:
\begin{itemize}
    \item $\forall x \in \mathcal{X}$,  
    \be\label{rkhs1}
    k_x= k(x, \cdot)\in \hh,
    \ee
    \item $\forall f \in \hh$, $\forall x \in \mathcal{X}$,  
    \be\label{rkhs2}
    f(x)=\scal{k_x}{f}_\hh.
    \ee
\end{itemize}
The latter condition is called the reproducing property of the kernel.  The following remark discusses the equivalence of the two above definitions.  

\br[Evaluation functionals and  reproducing kernel]
The existence of a reproducing kernel follows from the linearity and continuity of the evaluation functionals via the Riesz Representation Theorem.  
Conversely, the linearity and continuity of the evaluation functionals follow directly from Condition~\eqref{rkhs1} and the reproducing property.  
\er  

\br[Regularity properties of an RKHS]
It can be shown that the regularity properties of functions in an RKHS--e.g. measurability, continuity, or differentiability-- are inherited from the corresponding kernel.  
We do not develop this discussion here.  
\er

Another notion relevant for RKHS is that of a positive definite function, which leads to yet another equivalent definition.  

\paragraph{RKHS and positive definite kernels.}  

Recall that a function  $k:\mathcal{X} \times \mathcal{X} \to \mathbb{R}$ is called positive definite if, for all $x_1, \dots, x_N\in\mathcal{X}$, given $N\in\mathbb{N}$ and $c_1, \dots, c_N\in \mathbb{R}$,  
$$
\sum_{i,j=1}^N k(x_i,x_j)c_ic_j\ge 0.
$$  

It is easy to see that every reproducing kernel $k$ is symmetric and positive definite.  
Symmetry is straightforward, whereas positive definiteness can be proved using the reproducing property by noting that for all $x_i,x_j$ with $i,j=1, \dots, N$, we have $k(x_i, x_j) = \scal{k_{x_i}}{k_{x_j}}_{\hh}$, so that  
\begin{equation*}
    \sum_{i,j=1}^N k(x_i,x_j)c_ic_j = \scal{\sum_{i=1}^N c_i k_{x_i}}{\sum_{j=1}^N c_j k_{x_j}}_{\hh} =
    \nor{\sum_{i=1}^N c_i k_{x_i}}_\hh^2 \ge 0.
\end{equation*}

The converse of the above observation is known as the Moore-Aronszajn theorem.  

\br[Moore-Aronszajn theorem]\label{rm:MATheo}  
It can be shown that given a symmetric and positive definite kernel $k$, there exists a unique RKHS associated with it, for which $k$ is the reproducing kernel. \\  
The proof is constructive and is based on introducing the pre-Hilbert space  
$$
\hh_0=\{ f:\mathcal{X} \to \mathbb{R}~|~ \exists N\in \mathbb{N}, c_1, \dots, c_N\in \mathbb{R}, x_1, \dots, x_N\in \mathcal{X},  
~\text{s.t.}~f=\sum_{i=1}^N c_i k_{x_i}\}
$$  
endowed with the scalar product  
$$
\scal{f}{f'}_{\hh_0}= \sum_{i=1}^{N}\sum_{j=1}^{N'}k(x_i,x_j')c_ic_j', \quad  \forall f,f'\in \hh_0.
$$  
The inner product above can be shown to be well defined and independent of the choice of function representation.  
Then, the completion $\hh$ of $\hh_0$ is a Hilbert space, and it can be easily verified that it is indeed an RKHS with reproducing kernel $k$.  
\er

Finally, RKHSs are related to feature maps.  
\paragraph{RKHS and Feature Maps}  
Let $\mathcal{F}$ be a Hilbert space with inner product $\scal{\cdot}{\cdot}_{\mathcal{F}}$.  
A feature map is a function $\Phi: \mathcal{X} \to \mathcal{F}$ that embeds input points into $\mathcal{F}$.  

Every reproducing kernel $k: \mathcal{X} \times \mathcal{X} \to \mathbb{R}$ defines a canonical feature map by setting $\mathcal{F} = \hh$ and 
\be  
\Phi(x) = k_x, \quad \forall x \in \mathcal{X}.
\ee  
However, the same RKHS can define multiple feature maps.
For example, given any orthonormal basis $(a_j)_j$ of $\hh$, we can set $\mathcal{F} = \ell_2$ and define  
\be  
\Phi(x) = (a_j(x))_j, \quad \forall x \in \mathcal{X}.
\ee  

In turn, any feature map defines a corresponding RKHS. 
Indeed, given any $\Phi: \mathcal{X} \to \mathcal{F}$,  define  
\be  
\hh_{\Phi} = \{ f:\mathcal{X} \to \mathbb{R}~|~ \exists w \in \mathcal{F} \text{ s.t. } f(x) = \scal{w}{\Phi(x)}_{\mathcal{F}}, ~\forall x \in \mathcal{X} \}.
\ee  
The norm in $\hh_\Phi$ is given for all $f\in \hh_\Phi$ by
$$
\nor{f}_{\hh_\Phi}= \inf\{\nor{w}_{\mathcal{F}} ~|~ w\in \mathcal{F} \text{ s.t. } f(x) = \scal{w}{\Phi(x)}_{\mathcal{F}}, ~\forall x \in \mathcal{X} \}.
$$
This space consists of functions that are linear in the feature representation $\Phi(x)$.  
Since $\Phi$ might not be injective, the correspondence between $f \in \hh_\Phi$ and $w\in \mathcal{F}$ is not one-to-one, which justifies the infimum in the norm definition.

It is possible to show that the space $\hh_{\Phi}$ is an RKHS with reproducing kernel  
\be  
k(x, x') = \scal{\Phi(x)}{\Phi(x')}_{\mathcal{F}}.
\ee  
This proof, which is somewhat technical, is omitted.

The above discussion shows how a RKHS can be defined taking many different perspectives.
Before, discussing some examples we add a remark showing how  kernels can be combined to build new ones. 

\br[Closure Properties of RKHSs Under Sum and Product of Kernels]  

Given two reproducing kernels $k_1, k_2: \mathcal{X} \times \mathcal{X} \to \mathbb{R}$ with corresponding RKHSs $\hh_1$ and $\hh_2$, their sum and product define valid RKHSs. The sum  
\be  
k_+(x, x') = k_1(x, x') + k_2(x, x')
\ee  
is a reproducing kernel. The associated RKHS $\hh_+$ consists of functions $f = f_1 + f_2$ with $f_1 \in \hh_1$ and $f_2 \in \hh_2$. The norm in $\hh_+$ is given by  
\be  
\nor{f}_{\hh_+} = \inf \{ \sqrt{\nor{f_1}_{\hh_1}^2 + \nor{f_2}_{\hh_2}^2} ~|~ f = f_1 + f_2 \}.
\ee  

The pointwise product of kernels,
\[
k_\times(x, x') = k_1(x, x')\, k_2(x, x'),
\]
defines a reproducing kernel Hilbert space $\hh_\times$ with kernel $k_\times$. The space $\hh_\times$ contains functions that can be written as pointwise products $f_1 f_2$ with $f_1 \in \hh_1$ and $f_2 \in \hh_2$, and in fact includes the algebraic span of such products. In general, $\hh_\times$ is continuously embedded in the tensor product RKHS $\hh_1 \otimes \hh_2$.

\er

%
\paragraph{Examples of RKHSs}
We next discuss some basic examples of RKHSs and their corresponding reproducing kernels.
			
\bex[Linear kernel]
Let $(\XX, \scal{\cdot}{\cdot}_\XX)$ be a real separable Hilbert space. 
In particular, we could take $\XX=\R^d$. For all $x, x'\in\XX$, the linear kernel is $k(x, x') = \scal{x}{x'}_\XX$. Then 
for all $f\in \hh$, there exists a unique $w\in \XX$ such that 
$f(x)= \scal{w}{x}_\XX$  and  
$\nor{f}_\hh= \nor{w}_\XX$. 

To see this, note that from Equation~\eqref{rkhs1}, for all $x\in \XX$, $k_x\in \hh$. Since for the linear kernel, 
$k_x(\cdot)= \scal{x}{\cdot}_\XX\in \cL(\XX, \R)= \XX^*$, then $\XX^*=\hh_0 \subset \hh$.
Thus, we can write 
$$
\hh_0= \{f\in \hh~|~ \exists! w\in \XX ~\text{s.t.}~ f(\cdot)= \scal{w}{\cdot}_\XX\}.
$$
Moreover, for all $f,f'\in \hh_0$,
$$
\scal{f}{f'}_\hh= \scal{\scal{w}{\cdot}_\XX}{\scal{w'}{\cdot}_\XX}_\hh= \scal{w}{w'}_\XX,
$$
so that $\nor{f}_\hh= \nor{w}_\XX$.
Since $\hh_0$ is a closed subspace, we have the decomposition $\hh= \hh_0+\hh_0^\perp$. 
However, $\hh_0^\perp=\{0\}$, since by Equation~\eqref{rkhs2}, if $h\in \hh_0^\perp$, then for all $x\in \XX$,
$$
\scal{h}{\scal{x}{\cdot}_\XX}_\hh= h(x)= 0,
$$
which implies $h=0$. Thus, $\hh_0=\hh$, and the proof is finished.
\eex 
			
\bex[Gaussian kernel]\label{Gker}
Let $\XX=\R^d$ and, for $\gamma>0$, consider the Gaussian kernel $k(x, x')= e^{-\gamma \nor{x - x'}^2}$ for $x,x'\in \XX$. Then, for $f\in \hh\subset L^2(\R^d)$, the RKHS norm satisfies
\[
\nor{f}_\hh^2 \propto \int_{\R^d} |\widetilde f (\omega)|^2\, e^{\frac{\nor{\omega}^2}{\gamma}}\, d\omega,
\]
where $\widetilde f$ denotes the Fourier transform of $f$. In particular, functions in $\hh$ must have rapidly decaying Fourier transforms.
\eex 

\bex[Random features and integral representation kernel] 
Let $(\cB, \pi)$ be a probability space and define $L^2_\pi=\{g:\cB\to \R ~|~\nor{g}^2_{\pi}=\int |g(\beta)|^2d\pi(\beta)<\infty\}$. Let  
$\psi:\cB\times \X\to \R$ be a measurable function such that for almost all $x\in \X$, $\psi(\cdot, x)\in L^2_\pi$. Then, for all $x,x'\in \X$, define  
\be\label{IntRF}
k(x,x')= \int \psi(\beta, x)\psi(\beta, x')d\pi(\beta).
\ee
Next, consider $(\beta_i)_{i=1}^M\sim \pi^M$ and define  
$$
k_M(x,x')= \frac 1 M \sum_{i=1}^M \psi(\beta_i, x)\psi(\beta_i, x').
$$
The kernel $k_M$ is called the random features kernel.
\eex


%

\subsection{Regularized ERM in RKHS}

Let $\hh$ be a RKHS. Recalling the definition of the empirical risk~\eqref{emp_risk}, for each $\lambda>0$,  define the regularized empirical risk $\wh L _\la:\hh\to [0, \infty)$, such that for all $f\in \hh$,  
\be\label{RER}
\wh L_\la (f)= \wh L (f)+\la \nor{f}^2_\hh.
\ee
Then, consider the regularized empirical risk minimization problem,  
\be\label{RERM}
\min_{f\in \hh} \wh L_\la (f).
\ee
The functional $\wh L_\la$ is also  called the  Tikhonov functional; $\lambda$ is the regularization parameter, and $\nor{\cdot}^2_\hh$ is the regularizer.
If for all $y\in \Y$, $\ell(y, \cdot)$ is continuous and convex, then for all $\lambda>0$, $\wh L_\la$ is continuous, strictly convex, and coercive, so that there exists a unique minimizer $\wh f _\la\in \hh$. Next, we discuss how it can be computed considering the  square loss. 

\paragraph{Kernel ridge regression.}  
Let $\ell(y,y')= (y-y')^2$, for all $y,y'\in \R$. Then, the regularized ERM problem in RKHS is called kernel ridge regression (KRR). If $\X=\R^d$ and $k$ is the linear kernel, the method is called ridge regression (RR). It is useful to rewrite problem~\eqref{RERM} to highlight its connection to linear inverse problems. Towards this end, we introduce the sampling and extension operators defined by the kernel and the data.

\paragraph{Sampling and extension operators.}  
Endow $\R^n$ with the inner product  
\[
\scal{a}{a'}_n=\frac 1 n \sum_{i=1}^n a_i a_i', \quad \text{for } a, a' \in \R^n.
\]
Given an RKHS $\hh\subset \R^\XX$ with reproducing kernel $k$ and a set of $n$ input points $x_1, \dots, x_n\in \X$,  
define $\wh S: \hh \to \R^n$  for all $f\in \hh$ by  
\be\label{samp_op}
\wh S f = \left (\scal{f}{k_{x_1}}_\hh, \dots, \scal{f}{k_{x_n}}_\hh\right).
\ee  
Note that $\wh S$ is linear and bounded. Moreover, the adjoint $\wh S^*: \R^n\to \hh$ satisfies  for $c=(c_1, \dots, c_n)\in \R^n.$
\[
\wh S^* c = \frac 1 n \sum_{i=1}^n c_i k_{x_i}.
\]
We refer to $\wh S$ and $\wh S^*$ as the sampling and extension operators, respectively. The former name follows from the fact that  
\[
\wh S f = (f(x_1), \dots, f(x_n)).
\]
The latter is explained by considering $c= ({g}(x_1), \dots, {g}(x_n))$, the values of a function ${g}$ at $x_1, \dots, x_n$. Then, for all $x\in \XX$,  
\[
(\wh S^* c)(x)= \sum_{i=1}^n {g}(x_i) k(x_i, x),
\]
which can be interpreted as extending the values of ${g}$ to any other point through an averaging process performed via the kernel.

\paragraph{KRR and inverse problems.}
Given a training set of $n$ points, and a kernel $k$, let $\wh y= (y_1, \dots, y_n)\in \R^n$. Then for all $\la>0$, and $f\in \hh$
 \be\label{tik}
\wh L_\la (f) = \nor{\wh S f - \wh y}_n^2+\la \nor{f}^2_\hh.
\ee
The minimization of the above functional is the regularized least squares problem associated to the linear inverse problem
$$
\wh S f = \wh y.
$$

Then, taking the functional derivative w.r.t. $f$ in~\eqref{tik} and setting it to zero we get  that
 \be\label{tiksol}
\wh f _\la = (\wh S^* \wh S +\la I )^{-1}\wh S^* \wh y.
\ee
Next we show how the above  expression directly leads to computable quantities.

\paragraph{Representer theorem for KRR.}  
The solution $\wh f_\la$ admits an alternative representation,  
\be\label{KRR_repre1}
\wh f _\la =\wh S^* (\wh S \wh S^* +\la I )^{-1} \wh y.
\ee
The above expression can be derived in multiple ways, such as by direct manipulation using the Woodbury matrix identity or the singular value decomposition of $\wh S$. This derivation is left as an exercise. Here, we note that Equation~\eqref{KRR_repre1} can be further developed to show that for all $x\in \X$,  
\be\label{KRR_repre2}
\wh f_\la (x)= \sum_{i=1}^n k(x,x_i)\wh c_i,\quad \quad \wh c= (\wh K+n \la I)^{-1}\wh y,
\ee
where $\wh c \in \R^n$,  $\wh K = n \wh S\wh S^*$ is called the empirical kernel matrix and is symmetric, positive semi-definite, and such that $\wh K _{ij}=k(x_i, x_j)$ for all $i,j=1, \dots, n$.  

To derive~\eqref{KRR_repre2}, note that from Equation~\eqref{KRR_repre1}, $\wh f_\la = \wh S^* \wh c = \frac 1 n  \sum_{i=1}^nk_{x_i}\wh c_i$, where $\wh c=(\wh c_1, \dots, \wh c_n)\in \R^n$ satisfies  
$$
\wh c =(\wh S \wh S^* +\la I )^{-1} \wh y= \left (\frac{\wh K}{n} +\la I \right )^{-1} \wh y.
$$
Equation~\eqref{KRR_repre2} then follows by factoring out $n$.   The above expression shows that the kernel ridge regression (KRR) solution can be computed by solving a finite-dimensional linear system, even though the corresponding RKHS is infinite-dimensional. While feasible the above computations can become cumbersome for large kernel matrices. In the next section, we discuss how efficiency can be improved through approximate computations.

%
%
			
\subsection{Nystr\"om approximation}  
For $M\le n$, let $\{\widetilde x_1 , \dots, \widetilde x_M\}\subset \{ x_1 , \dots,  x_n\}$. The points $(\widetilde x_j)_{j=1}^M$ are called Nystr\"om centers or inducing points. Let $\widetilde Z: \hh \to \R^M$  
be such that for all $f\in \hh$,  
$$
\widetilde Z f =
\left (\scal{f}{k_{\widetilde x_1}}_\hh, \dots, \scal{f}{k_{\widetilde x_m}}_\hh\right).
$$  
Then, $\widetilde Z$ is linear and bounded. The adjoint $\widetilde Z^*: \R^M\to \hh$ is such that  
for $a=(a_1, \dots, a_M)\in \R^M$,  
$$
\widetilde Z^* a = \sum_{i=1}^M k_{\widetilde x_i}a_i.
$$  
Let $\widetilde \hh$ be the subspace of $\hh$ defined as  
$$
\widetilde \hh= \{f\in \hh~|~f= \widetilde Z^* a, \quad a\in \R^M\}.
$$  
Consider the regularized ERM problem on $\widetilde \hh$, given by  
\be\label{NERM}
\min_{f\in \widetilde \hh} \wh L _\la(f).
\ee  
We refer to this problem as the Nystr\"om KRR. Note that this minimization problem is defined by a strongly convex and coercive functional, so there is a unique solution $\widetilde f_\la \in \widetilde \hh$ such that  
$$
\wh L _\la(\widetilde f_\la)= \min_{f\in \widetilde \hh} \wh L _\la(f).
$$  
For  the square loss, we have that for all $f\in \widetilde\hh$,  
$$
\wh L _\la(f) = \nor{\wh S \widetilde Z^*a- \wh y }_n^2+\la \scal{a}{\widetilde Z \widetilde Z ^* a}_{\R^M},
$$  
so that problem~\eqref{NERM} is equivalent to  
$$
\min_{a\in \R^M} \nor{\wh S \widetilde Z^*a- \wh y }_n^2+\la \scal{a}{\widetilde Z \widetilde Z ^*a}_{\R^M}. 
$$  
By a direct computation, the solution of the above problem is  
\be\label{nrepre2}
\wt a = (\kmn\knm + \lambda n  \kmm)^{-1} \kmn \wh y,
\ee  
where $\knm = \wh S \wt Z^*$ is the matrix with entries $(\knm)_{ij}= k(x_i, \wt x_j)$ for $i=1, \dots, n$, $j=1,\dots, M$, and $\kmm = \wt Z\wt Z^*$ is the matrix with entries $(\kmm)_{ij}= k(\wt{x}_i, \wt x_j)$ for $i,j=1,\dots,M$.  
It is easy to see that for all $x\in \X$,  
\be\label{nrepre1}
\wt f_\la (x)= \sum_{i=1}^M k(\wt x_i,x)\wt a_i.
\ee

\begin{figure}
	\centering
	\begin{tikzpicture}
	\tikzset{
		mymat/.style={
			matrix of nodes,
			ampersand replacement=\&,
			nodes in empty cells,
			anchor=north, 
			column sep=0mm, row sep=0mm, nodes={
				anchor=center,
				inner sep=0pt,
				outer sep=0pt,
			}
		},
	}
	\node[mymat, draw=bluee, ultra thick, dashed, nodes={minimum size=2em}] (M) at (0, 0)
	{
		\& \& \& \\
		\& \& \& \\
		\& \& \& \\
		\& \& \& \\
	};
	\node[] at (M-2-2.south east) (k) {$\widehat{K}$};
	\node[fit=(M-1-1)(M-4-1), fill=bluee, fill opacity=0.4]{};
	\node[fit=(M-1-1), fill=malgared, fill opacity=0.7] { };
	\node[] at (M-1-1) {$\kmm$};
	\draw[-] (k.south west)--(k.north east);  
	
	\draw [decorate, decoration = {brace}, ultra thick] ([xshift=-7pt, yshift=-3pt]M-4-1.south west) -- node[left=2pt] {$\knm$} ([xshift=-7pt, yshift=3pt]M-1-1.north west);
\end{tikzpicture}
	\caption{The Nystr\"om approximation greatly reduces the size of the kernel matrices involved in solving the learning problem. Instead of $\widehat{K}$, only $\knm$ and $\kmm$ are needed.}
	\label{fig:colsubsampling}
\end{figure}
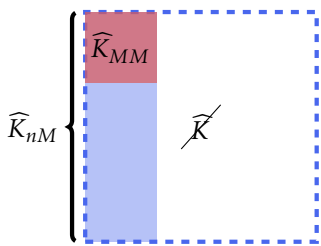

\br[Computational costs]
Equations~\eqref{nrepre2} and~\eqref{nrepre1} show that an efficient approximate KRR solution can be computed using a small number of Nystr\"om centers. Indeed, the cost for computing the exact KRR solution is $O(n^3)$ in time and $O(n^2)$ in space. The cost for computing the Nystr\"om KRR solution is 
$O(nM^2+M^3)$ in time and $O(nM)$ in space.
\er

\br[Column subsampling]
As already, via the representer theorem, KRR reduces to solving a linear system,
$
(\wh K +\la n I)c= \wh y,
$
which can be seen as a regularized version of the linear system,
$
\wh K c= \wh y. 
$
Analogously, Nystr\"om approximation can be seen to correspond to a regularized version of linear system
$
\wh K_{nM} c= \wh y. 
$
From this perspective Nystr\"om approximation corresponds to so called column subsampling, a technique from randomized numerical linear algebra. 
See Figure~\ref{fig:colsubsampling} for a pictorial representation.
\er
\br[Regularization by projection] 
Nystr\"om approximations can also be related to so called projection regularization methods. 
These methods naturally arise when the discretization of possible infinite dimensional problems need be considered.  Here,  the idea is to consider a subspace $\wt \hh$ and the corresponding orthogonal projection 
$\wt P:\hh\to \wt \hh$. Then, the following regularized problem
$$
\min_{f\in \hh} \nor{\wh S \wt P f - \wh y}_n^2+\la \nor{f}^2_\hh, 
$$
 can be shown to be equivalent to Problem~\eqref{NERM}. This perspective suggests more general ways to choose a subspace $\wt \hh$ than the one discussed before.
\er
\br[Approaches to select the Nystr\"om centers]
The Nystr\"om centers are often sampled at random from the training set. 
The simplest  idea is to  sample uniformly without replacement, but other strategies can be considered. Deterministic choices are also possible and are closely related to quadrature methods. 
\er

\subsection{Bibliography}
A classic reference for  reproducing kernel hilbert spaces is~\cite{aronszajn_1950} and the study of regularized  learning problem in  RKHS  originate in a number of foundational contributions in the '60s see e.g. ~\cite{wahba_book_1990} and references therein. 
Ridge regression was proposed in \cite{HoerlKennard1970} in the context of statistics, and is  called  Tikhonov regularization in inverse problems \cite{TikhonovArsenin1977,engl_regularization_2000}. It is a form of empirical risk minimization in statistical learning theory \cite{vapnik_slt_1998}. 
The formulation in terms of  sampling operators was introduced in \cite{devito_invprob_2005} and \cite{SmaleZhou2004}. 
The Nystr{\"o}m approximation was introduced in~\cite{williams_2000} and~\cite{smola_sparse_2000} under the name of sparse greedy approximation, and its statistical learning properties were proved in~\cite{rudi_less_2015} and \cite{rudi_falkon_2017} who also introduced a computationally efficient iterative algorithm, with connections to randomized linear algebra~\cite{martinsson_tropp_2020}. The use 
of 
General projection regularization was first considered in~\cite{engl_regularization_2000}.
\newpage

\section{Learning operators}\label{sec:op}

Instead of $\mathcal{Y} = \mathbb{R}$, consider $(\mathcal{Y}, \scal{\cdot}{\cdot}_\mathcal{Y})$ as a real separable Hilbert space. The statistical  learning problem naturally extends to this setting. Moreover, the approach based on kernels also generalizes provided a suitable notion of vector valued reproducing kernel Hilbert spaces is introduced. 

\subsection{Vector valued  statistical learning}
Let $(\mathcal{X}, \mathcal{A})$ be a measure space with $\sigma$-algebra $\mathcal{A}$,  
and let $(\mathcal{Y}, \scal{\cdot}{\cdot}_\mathcal{Y})$ be a real separable Hilbert space with its corresponding Borel $\sigma$-algebra $\mathcal{B}_\mathcal{Y}$.  
Let $(X,Y)$ be a pair of random variables taking values in $(\mathcal{X} \times \mathcal{Y})$ with law $P$.  
Let $\ell: \mathcal{Y} \times \mathcal{Y} \to [0,\infty)$ be a given measurable function.   Consider ${\mathcal{M}} (\mathcal{X}, \mathcal{Y}) \subset \mathcal{Y}^{\mathcal{X}}$, the space of all measurable functions from $\mathcal{X}$ to $\mathcal{Y}$.  
Define $L:{\mathcal{M}} (\mathcal{X}, \mathcal{Y})\to [0,\infty)$ as
$$
L(f) = \E[\ell(Y,f(X))], \quad \forall f\in {\mathcal{M}} (\mathcal{X}, \mathcal{Y}).
$$
The problem of learning is to solve 
\be\label{hilb_prob}
\min_{f\in \mathcal{M} (\mathcal{X}, \mathcal{Y})} L(f),
\ee
where $P$ is only known through a sample $(x_i,y_i)_{i=1}^n$ of $n$ independent and identically distributed copies of $(X,Y)$.  Our primary example for the loss function $\ell$ is the squared loss,
$$
\ell(y,y')= \nor{y-y'}_{\mathcal{Y}}^2, \quad \forall y,y'\in \mathcal{Y}.
$$
The case where $\mathcal{Y} = \mathbb{R}^T$ is a special case of this formulation.

\subsection{Vector valued RKHS} 

The definition of  vector valued reproducing kernel Hilbert space (vvRKHS) extend naturally to Hilbert spaces valued functions.  In the following, we let $(\mathcal{Y}, \scal{\cdot}{\cdot}_\mathcal{Y})$  be a real separable Hilbert space. 

\paragraph{Evaluation functionals and RKHS.}  
Let $\mathcal{X}$ be a set. A vector valued reproducing kernel Hilbert space  $\hh\subset \mathcal{Y}^\mathcal{X}$ is a Hilbert space with inner product $\scal{\cdot}{\cdot}_\hh$, such that for all $x \in \mathcal{X}$, the evaluation operators $e_x: \hh\to \mathcal{Y}$ defined for all $f\in \hh$ by  
\be\label{ev1}
e_x(f)= f(x)
\ee  
are linear and continuous. Note that this implies, in particular, that  for all $x\in \XX$ and $f\in \hh$
$$
\nor{f(x)}_{\mathcal Y}\lesssim \nor{f}_\hh.
$$  
Similarly, an equivalent definition can be given in terms of  reproducing kernels. However, for vector valued RKHS, reproducing kernels are  operator valued.
			
\paragraph{RKHS and reproducing kernels.}  
A vector valued reproducing kernel Hilbert space  $\hh\subset \mathbb{R}^\mathcal{X}$ is a Hilbert space with inner product $\scal{\cdot}{\cdot}_\hh$,  such that there exists $\Gamma:\X\times \X\to {\mathcal L} (\Y)$, called the operator valued reproducing kernel  satisfying the following properties:
\bi
\item $\forall x \in \X$,$\forall y \in \Y$,
\be\label{vvrkhs1}
\Gamma(x, \cdot)y\in \hh,
\ee
\item $\forall f \in \hh$, $\forall x \in \X$, $\forall y \in \Y$, 
\be\label{vvrkhs2}
\scal{\Gamma(x, \cdot)y}{f}_\hh=\scal{y}{f(x)}_\Y.
\ee
\ei 
The latter condition is  analogous to the reproducing property in the scalar valued setting. 
\br[Evaluation operator]
In view of the above definition, we  define the linear operator $\Gamma_x: {\mathcal Y}\to \hh$ such that 
$\Gamma_x= \Gamma(x, \cdot)$. Then, the corresponding adjoint operator $\Gamma_x^*: \hh\to {\mathcal Y}$ satisfies
$$
\Gamma_x^* f= f(x)
$$
by the reproducing property. Hence, $\Gamma_x^*$ provides a representation of the evaluation operator in Equation~\eqref{ev1}.
\er

As in the scalar case, the two definitions above can once again be related through an extension of the Moore--Aronszajn theorem and an application of the Riesz representation theorem. Similarly, vvRKHSs can be characterized in terms of positive definite functions and feature maps, but we omit these developments here. Instead, we provide some relevant examples of operator-valued kernels.

\bex[Operator valued linear  kernel]
Let $(\X, \scal{\cdot}{\cdot}_\X)$ be a {real separable} Hilbert space and let  $\Gamma(x, x')= \scal{x}{x'}_\X I_\Y$, for $x,x'\in \X$, where $I_\Y{:}\Y\to \Y$ is the identity operator.
 Then for all $f\in \hh$, there exists a unique $W\in \cL_2(\XX,\Y)$ such that for all $x\in\X$, $f(x)= Wx$ and  $\nor{f}_\hh= \nor{W}_{\cL^2(\XX,\Y)}$.
\eex 
\bex[Separable kernels I]
Let $\X$ be a set and $\hh_k\subset \R^\X$ a RKHS with corresponding scalar valued reproducing kernel $k:\X\times \X\to \R$. 
Let $\Gamma(x, x')= k(x,x') I_\Y$, for $x,x'\in \X$, where $I_\Y{:}\Y\to \Y$ is the identity operator.
 Then for all $f\in \hh$, there exists a unique $W\in {\mathcal L}_2(\hh_k,\Y)$ such that for all $x\in\X$, $f(x)= Wk_x$ and  $\nor{f}_\hh= \nor{W}_{\cL^2(\hh_k,\Y)}$.
We call these kernels separable, since the contribution of inputs and output space to the operator valued kernel is factorized.
\eex 
\bex[Separable kernels II]
The above examples can be further developed considering operator valued kernels of the form 
 $\Gamma(x, x')= k(x,x') A$, for $x,x'\in \X$, where $A:\Y\to \Y$ is a positive semi-definite operator.
\eex 
			
\bex[Integral representation and random features for  operator valued kernels] 
Let $(\cB, \pi)$ be a probability space and $L^2_\pi(\cB,\Y)=\{g:\cB\to \Y ~|~\nor{g}^2_{\pi}=\int |g(\beta)|^2d\pi(\beta)<\infty\}$. Then, let 
$\psi:\cB\times \X\to \Y$ a measurable function such that for almost all $x\in \X$, $\psi(\cdot, x)\in L^2_\pi(\cB,\Y)$. Then, for all $x,x'\in \X$, let 
$$
\Gamma(x,x')= \int \psi(\beta, x)\otimes\psi(\beta, x')d\pi(\beta).
$$
Next, consider $(\beta_i)_{i=1}^M\sim \pi^M$ and let 
$$
\Gamma_M(x,x')= \frac 1 M \sum_{i=1}^M \psi(\beta_i, x)\otimes\psi(\beta_i, x').
$$
The latter kernel is called a operator valued  random features  kernel.
\eex 
Next, we discuss ERM in a vvRKHS.

\subsection{Regularized ERM in vvRKHS}
The (regularized) ERM approach seamlessly extends to functions with values in a Hilbert space. 

\paragraph{Vector valued Kernel ridge regression.}
Let $\ell(y,y') = \nor{y - y'}_\Y^2$, for $y, y' \in \Y$, and consider the extension of Equations~\eqref{RER} and~\eqref{RERM}. In this case, we call the regularized ERM in RKHS \emph{vector-valued kernel ridge regression} (vvKRR). We now develop a discussion analogous to the one in the scalar setting.

\paragraph{Sampling and extension operators.} {Given an operator valued reproducing kernel $\Gamma$} with RKHS $\hh$, for all $x\in \X$, let $\Gamma_x:\Y\to \hh$ be the linear bounded operator  such that 
$\Gamma_x^*:\hh\to \Y$ is given by  $\Gamma_x^* f= f(x)$ for $f\in \hh$, $x\in \X$. 
These operators are well  defined in view of Conditions~\eqref{vvrkhs1},~\eqref{vvrkhs2}.
Also, note that for all $x,x'\in \X$, $\Gamma_x^*\Gamma_{x'}= \Gamma(x,x')$.
Let $\Y^n = \oplus _{i=1}^n\Y$,  with the normalized inner product $\scal{a}{a'}_{\Y^n}=\frac 1 n \sum_{i=1}^n \scal{a_i}{a_i'}_\Y$ for $a, a\in \Y^n$.
{Given a training set, let  $x_1, \dots, x_n\in \X$ be the corresponding input points.} 
{Define the  sampling operator for  vvRKH spaces as } $\wh S: \hh \to \Y^n$, such that for all $f\in \hh$,
$$
\wh S f = \left (\Gamma^*_{x_1}f, \dots, \Gamma^*_{x_n}f\right).
$$ 
Then, $\wh S$ is linear and bounded.  Moreover, the corresponding extension operator is the adjoint  $\wh S^*: \Y^n\to \hh$  such that 
for $a=(a_1, \dots, a_n)\in \Y^n$
$$
\wh S^* a = \frac 1 n \sum_{i=1}^n \Gamma_{x_i}a_i.
$$

\paragraph{vvKRR solution and representer theorem.}

{Given a training set, and an operator valued kernel $\Gamma$, let $\wh y= (y_1, \dots, y_n)\in \Y^n$. Then for all $\la>0$ 
$$
\wh L_\la (f) = \nor{\wh S f - \wh y}_{\Y^n}^2+\la \nor{f}^2_\hh, \quad f\in \hh.
$$
The same computations done for scalar functions show that 
$$
\wh f _\la = (\wh S^* \wh S +\la I{_\Y} )^{-1}\wh S^* \wh y= \wh S^* (\wh S \wh S^* +\la I{_\Y} )^{-1} \wh y.
$$
Further,  we can also write   for all $x\in \X$, 
\be\label{vvKRR_repre}
\wh f_\la (x)= \sum_{i=1}^n \Gamma(x,x_i)\wh c_i,\quad \quad \wh c= (\wh \Gamma+\la n  I{_{\Y}})^{-1}\wh y,
\ee
where $\wh c = (\wh c_1, \dots, \wh c_n)\in \Y^n$, $\wh \Gamma = n \wh S\wh S^*$ is the corresponding empirical kernel operator. 
\bex[Computations with operator valued linear kernels]
Let $(\XX, \scal{\cdot}{\cdot}_\XX)$ be a real {separable} Hilbert space and $\hh\subset \Y^\XX$ a vvRKHS with reproducing kernel $\Gamma$.  If $\Gamma=\scal{\cdot}{\cdot}_\XX I_\Y$, then the vvKRR problem can be written as, 
$$
\min_{W\in \cL^2(\XX, \Y)} \frac 1 n \nor{\wh Y - \wh X W^*}_{\cL^2(\Y,\R^n)}^2+\la \nor{W}^2_{\cL^2(\X,\Y)},
$$
where $\wh Y:\Y\to \R^n $ and $\wh X:\XX \to \R^n $
and the corresponding solution is 
$$
\wh W =  \wh Y^*\wh X (\wh X^* \wh X +\la I n)^{-1} 
$$
\eex

\br[Discretization]
The above computations in general need some discretization to be performed. 
For example, considering $\X_d\subset \X$, $\Y_T\subset\Y$ subspaces of dimensions $d$ and  $T$, respectively. 
\er

\br[Finite dimensional output spaces]
Assume $(\X, \scal{\cdot}{\cdot}_\hh)$ and $\Y= \R^T$. In particular, consider the operator (matrix) valued linear kernel $\Gamma(x,x')=\scal{x}{x'}_\X I_\Y$.  In this case, the empirical kernel operator can be identified with a $nT\times nT$ block diagonal matrix. Note that an analogous observation holds for separable kernels.
\er

\subsection{Bibliography}
Vector valued RKHSs were   introduced in ~\cite{schwartz_vec_1964}
and we refer to  \cite{carmeli_2006} for more recent results and further references. Their potential use in machine learning was highlighted in  \cite{micchelli_2005}, while  \cite{caponnetto_optimal_2007} proved the first results on vector valued KRR. We refer to 
\cite{Kov2024} for more recent results and further references in the context of operator learning. 

\newpage
\section{Learning dynamical systems}

We next describe how discrete time stochastic dynamical systems can be learned using ideas from operator learning, by leveraging the Koopman operator theory. 

\subsection{Dynamical systems and Markov processes}

The term {dynamical system} broadly refers to a quantity that evolves over time. For quantities represented by  vectors in  \(\mathcal{X} \subseteq \mathbb{R}^d\), a discrete-time evolution can be described by the iterative map  
\[
x_{t+1} = f(x_t),
\]  
for a given initial condition \(x_0\). Here, \(\mathcal{X}\) is called the { state space}, and its elements are called {states}, while the function \(f:\mathcal{X} \to \mathcal{X}\) is called the {evolution function}. Such dynamical systems are called {autonomous} because the evolution function does not  depend on time.

In many situations, it is useful to consider {stochastic dynamics}, given by  
\be\label{sds}
x_{t+1}= f(x_t, \omega_t),
\ee
where the initial state \(x_0\) is drawn from a given initial distribution \(\rho_0\) on $\XX$.  
Here, \((\omega_t)_{t\in \mathbb{N}}\) are i.i.d. samples in some probability space \((\Omega, \mathcal A, \mathbb P)\), and encode the stochastic nature of the evolution. Stochasticity may arise as a perturbation to an underlying deterministic dynamics, or be intrinsic to the evolution itself.
Stochastic dynamical systems can  be equivalently described in terms of Markov processes, as described next. 

\paragraph{Markov processes.}
Let \((X_t)_{t\in \mathbb{N}}\) be a stochastic process with values in a measurable space \((\mathcal{X}, {\mathcal A})\). Assume that for all measurable set \( A \in   \mathcal{A} \) and for every \( t\in \mathbb{N} \),
\be\label{MC}
\mathbb{P}(X_{t+1} \in A \mid X_t, \dots, X_1) = \mathbb{P}(X_{t+1} \in A \mid X_t).
\ee

Then, \((X_t)_{t\in \mathbb{N}}\) is called a {Markov process}. The Markov process is called {time-homogeneous} if the conditional probability in Equation~\eqref{MC} is the same for all \( t \in \mathbb{N} \).  In this case, there exists a transition kernel \( p: \mathcal{X} \times \mathcal{A} \to [0,1] \) such that for all \( t \in \mathbb{N} \),  
\[
p(x, A)= \mathbb{P}(X_{t+1} \in A \mid X_t = x),
\]
for all \( x \in \mathcal{X} \) and all \( A \in \mathcal{A} \). Conversely, given a transition kernel $p$ and an initial distribution $\rho_0$, a Markov process can be defined letting $X_0\sim \rho_0$ and for all $x\in \XX,  A\in \mathcal{A}$,
\[
\mathbb{P}(X_{t+1} \in A \mid X_t = x) = p(x, A).
\]

\br[Stochastic dynamical systems and Markov processes]
It can be shown that every stochastic dynamical system~\eqref{sds} defines a Markov process and conversely, any Markov process can be realized by a stochastic dynamical system~\eqref{sds}. 
Roughly speaking, the autonomous nature of the system and the i.i.d. nature of the samples in~\eqref{sds} translate into the Markovianity of the process.
\er
An important notion associated with a Markov process with transition kernel \( p \) is that of  
an {invariant measure}, which is a probability measure \(\pi\) on \(\mathcal{X} \) such that
\[
\pi(A) = \int_{\mathcal{X}} p(x, A) \, d\pi(x), \quad \forall A \in \mathcal{A}.
\]
In general, we cannot expect a Markov process to have an invariant measure, but existence is ensured in a number of relevant cases.
A sufficient condition is given in the following remark.

\br[Positive Harris recurrence and invariant measures]  
Let \( (X_t)_{t \in \mathbb{N}} \) be a Markov process with transition kernel \( p \). For  \( A \in \mathcal{A} \), define the return time to \( A \) as  
\[
\tau_A = \inf \{ t \geq 1 : X_t \in A \}.
\]
The process is called {Harris recurrent} if there exists a  $\sigma$-finite measure \( \mu \) on \( \X \) such that for all measurable \( A \subseteq \X \) with \( \mu(A) > 0 \) and all \( x \in \X \),
\[
\mathbb{P}_x(\tau_A < \infty) = 1.
\]
The process is positive Harris recurrent if it is Harris recurrent and there exists a set \( A \) with \( \mu(A) > 0 \) such that
\[
\mathbb{E}_x[\tau_A] < \infty, \quad \forall x \in A.
\]
Under these conditions, the process admits a unique invariant probability measure \( \pi \).
\er

Another important notion is {time reversibility}, characterized by the so-called {detailed balance condition}
\be\label{DB}
p(x, dx') d\pi(x) = p(x', dx) d\pi(x').
\ee
More precisely, the above shorthand notation means that, for all measurable sets \( A, B \subseteq \mathcal{X} \),
\[
\int_A p(x, B) d\pi(x) = \int_B p(x', A) d\pi(x').
\]
Intuitively, this  means that the Markov process behaves the same way forward and backward in time.

Finally, it is useful to recall that linear dynamical systems have special properties that greatly simplify their study.

\br[Linear dynamical system]\label{lin_syst}
Let \( A \in \mathbb{R}^{d\times d} \), and consider the dynamical system  
\be\label{lds}
x_{t+1}= A x_t, 
\ee
for some initial condition \( x_0 \). The study of the system dynamics can be greatly simplified if \( A \) admits a spectral decomposition
\be\label{spectral}
A= \sum_{j=1}^d \lambda_j \psi_j \otimes \psi_j,
\ee
for some suitable eigenvalues \( \lambda_1, \dots, \lambda_d \in \mathbb{R} \) and eigenvectors \( \psi_1, \dots, \psi_d \in \mathcal{X} \). This is the case, for instance, if \( A \) is symmetric or normal. Using the spectral decomposition~\eqref{spectral}, the system~\eqref{lds} can be written as  
\[
x_t = A^t x_0 =
\sum_{j=1}^d \lambda_j^t \scal{\psi_j}{x_0} \psi_j.
\]
This expression shows that the evolution of \( x_t \) is governed by the eigenvalues \( \lambda_j \), with modes associated with \( |\lambda_j| > 1 \) growing and those with \( |\lambda_j| < 1 \) decaying over time.
\er  

In the following, we focus on nonlinear systems and discuss how the so-called Koopman operator theory extends ideas from linear systems.


%

\subsection{Koopman operator theory}		
Consider a Markov process with transition kernel \( p \) and invariant measure \( \pi \).
Let \( L^2_\pi= L^2_\pi(\mathcal{X}, \mathbb{R}) = \{g:\mathcal{X} \to \mathbb{R} \mid \nor{g}_\pi=\int |g|^2 d\pi < \infty\} \). Define the operator \( A_\pi: L^2_\pi \to L^2_\pi \) as
\[
A_\pi g(x) = \mathbb{E}[g(X_{t+1}) \mid X_t = x],
\]
for all \( g \in L^2_\pi \) and almost all \( x \in \mathcal{X} \).  
Equivalently, writing the expectation explicitly,
\[
A_\pi g(x) = \int_{\mathcal{X}} g(x') p(x, dx').
\]
The operator \( A_\pi \) is called the Koopman operator or Markov  operator. It is linear and bounded with \( \|A_{\pi}\| \leq 1 \). Moreover, for all \( \tau \in \mathbb{N} \), it holds that
\[
A^\tau_\pi g(x) = \mathbb{E}[g(X_{t+\tau}) \mid X_t = x].
\]

The interpretation is that any \( g \in L^2_\pi \) represents an observable of the system. Its evolution over time follows a linear transformation given by the Koopman operator.

If the Markov process is time reversible, then the Koopman operator is self-adjoint, that is, \( A_\pi = A^*_\pi \).

\br[Self-adjointness of the Koopman operator]  
For all \( g, h \in L^2_\pi \), using the definition of the Koopman operator 
\[
\langle A_\pi g, h \rangle_{L^2_\pi} = \int_{\mathcal{X}} A_\pi g(x) h(x) d\pi(x)
= \int_{\mathcal{X}} h(x) \left (\int_{\mathcal{X}} g(x') p(x, dx')\right) d\pi(x).
\]
Swapping the integrals and using the detailed balance condition~\eqref{DB}, we get  
\[
\langle A_\pi g, h \rangle_{L^2_\pi} =  \int_{\mathcal{X}} g(x')\left ( \int_{\mathcal{X}} h(x) p(x', dx) \right)d\pi(x')
= \int_{\mathcal{X}} g(x') A_\pi h(x') d\pi(x')=\langle  g, A_\pi h \rangle_{L^2_\pi}
\]
which shows that \( A_\pi \) is self-adjoint.  
\er

If the transition kernel is absolutely continuous with respect to the Lebesgue measure, with a square integrable density, then the Koopman operator is also compact. 

\br[Compactness of the Koopman operator]  
Let \( q(x, x') \) be the density of the transition kernel with respect to the Lebesgue measure, i.e.,  
$
p(x, dx') = q(x, x') dx'.
$
Then, the Koopman operator  \( A_\pi \) is the integral operator given by  
\[
A_\pi g(x) = \int_{\mathcal{X}} q(x, x') g(x') d\pi(x').
\]
It is a standard fact in functional analysis that, if 
\[
\int_{\mathcal{X}} \int_{\mathcal{X}} q^2(x, x') d\pi(x) d\pi(x') < \infty,
\]
then the Koopman operator is a Hilbert-Schmidt operator and hence compact.  
\er

In the following, we assume the Koopman operator \( A_\pi \) to be self-adjoint and compact. Then,  \( A_\pi \) admits an orthonormal eigensystem \( (\lambda_i, \psi_i)_{i\in \N} \), with \( \lambda_i \geq 0 \) and \( \psi_i \in L^2_\pi \), satisfying  
\[
A_\pi \psi_i = \lambda_i \psi_i, \quad i = 1, 2, \dots,
\]
and by the spectral theorem,
\[
A_\pi = \sum_{i=1}^\infty \lambda_i \psi_i \otimes \psi_i.
\]
This expansion is known as the Koopman mode decomposition. The functions \( \psi_i \) are called Koopman modes and  represent spatial patterns of the system, while the eigenvalues \( \lambda_i \) characterize their temporal evolution. Note that, for $t\in \N$ and $g\in L_\pi^2$
\[
A^t_\pi g= \sum_i \lambda_i^t \langle g, \psi_i \rangle_{L^2_\pi}  \psi_i.
\]
This decomposition provides a spectral perspective on nonlinear dynamical systems, analogous to the spectral expansion of linear systems discussed earlier in Remark~\ref{lin_syst}.

The above discussion highlights how nonlinear dynamical systems can be characterized in terms of corresponding operators. In practice,  Koopman operators might be  hard to compute exactly, but empirical approximations can be derived based on observations from the corresponding dynamical systems. We next show how considering observables in an RKHS allows casting the estimation of the Koopman operator as a suitable operator learning problem.
\subsection{Learning the Koopman operator with kernels}

Let $\hh\subset \R^\X$ be a scalar reproducing kernel Hilbert space with reproducing kernel $k$. 
The idea is to consider observables in $\hh$ to estimate $A_\pi$ from data.
We next provide an intuition for why restricting observables to an RKHS enables efficient algorithm design.

\paragraph{Kernel Koopman regression.} 
The idea is to find $W\in {\mathcal L}(\hh)$ such that for any $f \in \hh$ and any $t\in \N$
$$
f(X_{t+1})\approx W f(X_t).
$$
A useful observation is that if $f\in \hh$, then $\forall W\in {\mathcal L}(\hh)$ and $\forall t\in \N$, by the reproducing property
$$
\E[(f(X_{t+1})- Wf(X_t))^2]= 
\E[\scal{f}{\phi(X_{t+1})-W^*\phi(X_{t})}_\hh^2], 
$$
where  $\phi(x) = k_x=k(x,\cdot)$ for all $x\in \XX$.
Then, if $(f_j)_j$ is an orthonormal basis of $\hh$, then for all $t\in \N$,
$$
\sum_j\E[(f_j(x_{t+1})- Wf_j(X_t))^2]= 
\sum_j
\E[\scal{f_j}{\phi(X_{t+1})-W^*\phi(X_{t})}_\hh^2]=
\E[ \nor{\phi(X_{t+1})-W^*\phi(X_{t})}^2_\hh].
$$
Finally, define the risk $L:\mathcal{L}(\hh)\to \mathbb{R}$, for all $W\in {\mathcal L}(\hh)$ by
\be\label{kooprisk}
L(W)= \E[ \nor{\phi(X_{t+1})-W^*\phi(X_{t})}^2_\hh]. 
\ee		
The above computations lead to several observations.
First, they suggest that considering an RKHS enables studying the evolution of the kernel rather than that of all observables.  
Second, given samples \( x_1, \dots, x_T \), we can estimate the expectation in~\eqref{kooprisk} by the empirical risk \( \wh L:\mathcal{L}(\hh)\to \mathbb{R}\) for all \( W\in {\mathcal L}(\hh) \) as
\be\label{koopempirical}
\wh L(W)= \frac 1 T  \sum _{t=0}^{T-1}\nor{\phi(x_{t+1})-W^*\phi(x_{t})}^2_\hh.
\ee	
Before proceeding further, we  discuss some implications of restricting observables to an RKHS.

\paragraph{RKHS restriction of the Koopman operator.}

Next, we develop the above dobservations,  discussing  how an operator $W\in {\mathcal L}(\hh)$ can approximate  the Koopman operator, which naturally belongs to ${\mathcal L}(L^2_\pi)$. We show that rather than the Koopman operator itself we are approximating its restriction to the RKHS. 

To show  this, we define a suitable  embedding  operator defined by the kernel and the invariant measure. Assume that there exists $\kappa > 0$ such that for all $x\in \XX$, 
\be\label{bound_ker}
k(x,x)\le  \kappa^2.
\ee
Define $S:\hh\to L^2_\pi$  by
\be\label{S}
Sf(x)= \scal{f}{k_x}_\hh,
\ee
for all $f\in \hh$ and  almost all $x\in \X$. Several observations can be made. First, the operator $S$ is linear and bounded by Assumption~\eqref{bound_ker}. Indeed, for all $f\in \hh$, we have
$$
\nor{S f}_\pi\le  \kappa  \nor{f}_\hh,
$$
by the reproducing property~\eqref{rkhs2} and Assumption~\eqref{bound_ker}. Second, $S$ is the embedding operator from $\hh$ to $L_\pi^2$, so that each function $f\in \hh$ is seen as an element $Sf\in L^2_\pi$, with its norm changing from that in $\hh$ to that in $L^2_\pi$.  Third, if the support $\XX_\pi$ of the invariant measure is strictly contained in $\XX$, then $S$ is not injective and acts as a restriction operator. From this latter perspective, it can be seen as a continuous analog of the sampling operator~\eqref{samp_op}.  Finally, it is easy to check that the operator $S$ is Hilbert-Schmidt.
Let $(f_j)_j$ be an orthonormal basis of $\hh$. Then 
\beas
\nor{S}_{\cL(\hh, L^2_\pi)}^2 &=&
\text{Tr}(S^* S)= 
\sum_{j=1}^\infty \scal{S^* Sf_j}{f_j}_\hh= 
\sum_{j=1}^\infty \scal{ Sf_j}{Sf_j}_{L_\pi^2}\\
&=&
\sum_{j=1}^\infty \int   |\scal{k_x}{f_j}_\hh|^2 d\pi(x)=
\int  \sum_{j=1}^\infty |\scal{k_x}{f_j}_\hh|^2 d\pi(x)\\
&=&\int k(x,x)d\pi(x)\le \kappa^2.
\eeas

Given the above premise, we derive an identity providing  insights into the nature of the approximation attainable by minimizing the risk~\eqref{kooprisk}, as well as  the empirical risk~\eqref{koopempirical}.  

Define $A_\hh:\hh\to L^2_\pi$  as $A_\hh=A_\pi S$, that is, the restriction of the Koopman operator to the RKHS.  Note that, $A_\hh  \in \cL^2(\hh, L^2_\pi)$ since 
the set of Hilbert-Schmidt operators forms a two-sided ideal in the algebra of bounded operators. 
Similarly, $SW\in \cL^2(\hh, L^2_\pi)$ for all $W\in \cL(\hh)$. Next we show  that 
\be\label{risk_decomposition}
L(W)= \nor{A_\hh- SW}_{\cL^2(\hh, L^2_\pi)}^2+ \sigma^2, 
\ee
for some suitable constant $\sigma^2$. The above expression shows that  when minimizing the  risk~\eqref{kooprisk} we are effectively finding an approximation to the Koopman operator restricted to the RKHS. 

To prove Equation~\eqref{risk_decomposition} we begin noting that for all $W\in \cL(\hh)$
\be\label{classic}
L(W)= 
\E[ \nor{\phi(X_{t+1})-W^*\phi(X_{t})}^2_\hh]= \E[ \nor{F_*(X_t)-W^*\phi(X_{t})}^2_\hh] +\sigma^2, 
\ee
where for almost all $x\in \X$
$$
F_*(x)= \E[ \phi(X_{t+1})~|~ X_t= x]
$$
and 
$$
\sigma^2= \E[\nor{\phi(X_{t+1})-F_*(X_t)}_\hh^2].
$$
Equation~\eqref{classic} is a classic result that can be easily checked developing the square and taking the expectation. We omit this calculation to note that for all $f\in \hh$, by linearity of the expectation
$$
\scal{F_*(x)}{f}_\hh= \E[ \scal{\phi(X_{t+1})}{f}_\hh~|~ X_t= x]=\E[ Sf(X_{t+1})~|~ X_t=x] = A_\pi S f(x),  
$$
and 
$$
\scal{W^*\Phi(x)}{f}_\hh= \scal{\Phi(x)}{Wf}_\hh= SWf(x)
$$
for almost all $x\in \X$. Moreover,  recall that, for any orthonormal basis $(f_j)_j\in \hh$ and  $B\in \cL_2(\hh, L^2_\pi)$, 
$$
\nor{B}_{\cL_2(\hh, L^2_\pi)} = \sum_j \nor{Bf_j}_{L^2_\pi}^2.
$$
Then, any orthonormal basis $(f_j)_j\in \hh$  the following equalities hold
\beas
\E[ \nor{F_*(X_t)-W^*\phi(X_{t})}^2_\hh]&=& \sum_{j}  \E[ (\scal{F_*(X_t)-W^*\phi(X_{t})}{f_j}_\hh)^2]\\
&=&\sum_{j}  \E[ (A_\pi S -SW)f_j(X_t))^2]\\
&=&   \sum_{j} \nor{(A_\pi S -SW)f_j}^2_{L^2_\pi}\\
&=& \nor{A_\pi S -SW}_{\cL_2(\hh, L^2_\pi)}^2
\eeas
Combining the above expression in Equation~\eqref{classic} leads to~\eqref{risk_decomposition}.
Provided with the  above discussion,  we next discuss the computation of the empirical approximation of the Koopman operator restricted to a RKHS.
%
%

\paragraph{KRR for the Koopman operator.}
Following the discussion in the previous section, given a sample trajectory \( x_1, \dots, x_T \) and an RKHS \( \hh \) with reproducing kernel \( k \),  consider the regularized empirical risk minimization problem
\be\label{kooprerm}
\min_{W\in \cL_2(\hh)} \frac 1 T \sum_{t=1}^T \|\phi(x_{t+1}) - W^* \phi(x_t)\|_\hh^2 + \lambda \|W\|^2_{\cL_2(\hh)}.
\ee

Note that, we further restricted the class of operator considering $\cL_2(\hh)$ rather than $\cL(\hh)$. As discussed in Section~\ref{sec:op}, this corresponds to a suitable operator valued reproducing kernel and allows the computation of the solution denoted by  \( \wh W_\lambda \). 

Let $\wh Y :\hh\to \R^T$ be such that $(\wh Y f)^t=\scal{k_{x_{t+1}}}{f}_\hh$,  and 
$\wh X :\hh\to \R^T$ be such that $(\wh X f)^t=\scal{k_{x_{t}}}{f}_\hh$, \pz{t=1, \dots, T}. 
Then,   from the optimality condition of problem~\eqref{kooprerm}   the following formulas can be derived,
$$
\wh W_\la= \wh Y^* \wh X(\wh X^*\wh X+T\la I )^{-1} =  \wh Y^* (\wh X\wh X^*+T\la I )^{-1}\wh X
$$
Moreover, for all $x \in \XX$ let 
\[\wh k(x)= (k(x_1, x), \dots, k(x_T, x))\]
and 
$$
\wh \alpha(x)=(\wh X\wh X^*+T\la I )^{-1}\wh{\kappa}(x)\in \R^T
$$
Note that, while $\wh W_\la^*$ is infinite dimensional, the coefficient vector  $\wh \alpha(x)$ is finite 
dimensional  for all $x\in \XX$, and further its computations involves only finite dimensional quantities.
Several observations a can be made. First,  it is possible to predict any observable given a number of measurements $f(x_1),\dots, f(x_{T-1})$. Indeed, for all $f\in \hh$, and $x\in \X$
\beas
\wh W _\la f(x)&=&\scal{\wh W _\la f}{k_x}_\hh \\
&=&\scal{f}{ \wh Y^* (\wh X\wh X^*+T\la I )^{-1}\wh X k_x}_\hh\\
&=&\scal{\wh Y f}{ (\wh X\wh X^*+T\la I )^{-1}\wh X k_x}_{\R^T}\\
&=& \sum_{t=1}^T f(x_{t+1})\alpha(x)_t.
\eeas
Note that in particular, we could consider $f(x)=x^j$ for $x= (x^1, \dots,x^d)$ to forecast  (the coordinate of) future states.
Second, it possible to show that empirical  Koopman modes can also be computed but the reasoning is more 
involved and is omitted here.

\subsection{Bibliography}
An introduction to several ideas related to  dynamical systems and their empirical estimation 
can be found in \cite{brukut22}. A standard reference for Markov processes is for example \cite{meytwe09}. 
Koopman operator theory is discussed in  \cite{giannakis_data_2019,mezic_koopman_2021,mezic_koopman_2021}.
The learning approach to dynamical systems  based on Koopman operators and reproducing kernel Hilbert spaces is discussed in  \cite{kostic_learning_2022}.


		
\section*{Acknowledgments}

First, I would like to thank Claudio Agostinelli for his patience and support. I am grateful to him and to Massimo Fornasier for inviting me to co-organize and teach at the "Machine Learning: From Data to Mathematical Understanding" school in Cetraro. This manuscript is the supporting material for my lectures there. Note that the emphasis here is on simplifying the exposition, and I have included only a few basic references rather than attempting to provide a comprehensive survey of all contributions and results.

 I would  like to thank Giacomo Meanti and Pietro Zerbetto, who contributed to these notes, and Oleksii Kachaiev, who proofread part of the manuscript. More generally, I would like to thank  Ernesto De Vito and all the many co-authors whose results are presented in these notes. I hope I have done justice to their work. 
 
I acknowledge the financial support of the European Commission (Horizon Europe grant ELIAS 101120237), the Ministry of Education, University and Research (FARE grant ML4IP R205T7J2KP), the European Research Council (grant SLING 819789), the US Air Force Office of Scientific Research (FA8655-22-1-7034), the Ministry of Education, University and Research (grant BAC FAIR PE00000013, funded by the EU – NGEU), and MIUR (PRIN 202244A7YL). This work represents only the views of the authors. The European Commission and the other organizations are not responsible for any use that may be made of the information it contains.
	
\appendix
\section{Basic facts and notation}
We  recall a few basic definitions useful in the following. 
\paragraph{Random variables and their laws.}
Let $(\Omega,\mathcal A, \mathbb P)$ a probability space with measure $\P$, and $(\Z, \mathcal B)$ a measure space.  Let $Z: \Omega \to \Z$ be a measurable function, then $Z$ is called a random variable. The law of $Z$ is the measure $P=P_Z$ on $\X$ given for almost all $A\in\mathcal A$ by
$$
P_Z(A)= \P (Z^{-1}(A)).
$$
The following notation is often used
$$
{\mathbb P}(Z\in B)= \P( \{\omega \in \Omega ~|~Z(\omega)\in B\}).
$$
for $B\in \mathcal B$. 
Let $\Y$ be a Hilbert space and $g:\Z\to \Y$ a measurable function.
$$
\E[g(Y)]=\int g(x)dP_Z(z)=\int g(Z(\omega))d\P(\omega).
$$
A stochastic process in $\X$ is a family of random variables $(Z_t)_{t\in {\mathcal T}}$, indexed over some set $\mathcal T$.

\paragraph{Operators and their norms.}

Let $\hh, \G$ be Hilbert spaces. Let $\cL(\hh, \G)$ be the space of linear bounded operators from $\hh$ to $\G$. If $\G=\R$ then $\cL(\hh, \R)$ is the space of linear continuous functionals on $\hh$.
Let $\cL(\hh)=\cL(\hh, \hh)$. The operator norm is defined for all $A\in \cL(\hh, \G)$ as 
$$
\nor{A}_{\cL(\hh, \G)}= \sup_{\nor{h}_\hh\le 1}\nor{Ah}_\G
$$

 The adjoint $A^*\in \cL(\G, \hh)$ is the unique linear bounded operator such that for all $h\in \hh, g\in \G$, 
$$
\scal{Ah}{g}_\G= \scal{h}{A^*g}_\hh.
$$
If $(e_j)_j\in \hh$ is an orthonormal basis, 
the trace of $A\in \cL(\hh)$ is defined 
$$
\text{Tr}(A)=\sum_{j}\scal{Ae_j}{e_j}_\hh 
$$
and is independent to the choice of the basis. 
 Let $\cL^2(\hh, \G)$ be the Hilbert space of Hilbert-Schmidt operators with inner product
$$
\scal{A}{B}_{\cL^2(\hh, \G)}= \text{Tr}(A^*B).
$$
The space of Hilbert Schmidt operators $\cL(\hh)$ is an ideal in $\cL(\hh, \G)$, that is,  if 
$A\in \cL_2(\hh), B\in \cL(\hh, \G)$ then $BA\in \cL_2(\hh, \G)$ and 
$$
\nor{BA}_{\cL^2(\hh, \G)}\le \nor{B}_{\cL(\hh, \G)}\nor{A}_{\cL^2(\hh)}.
$$

\paragraph{Optimization, coercivity and convexity.}
Let $\hh$ a Hilbert space and $F:\hh\to \R$ a convex functional. We say that $F$ is coercive if 
$$
\lim_{\nor{f}\to \infty} F(f)= \infty.
$$
If $F$ is continuous, or even just lower semi-continuous, and coercive then it has a non empty set of minimizers. If $F$ is strictly convex then there is a unique minimizer.

\bibliographystyle{plain}
\bibliography{biblio}

\begin{thebibliography}{10}

\bibitem{aronszajn_1950}
N.~Aronszajn.
\newblock Theory of reproducing kernels.
\newblock {\em Transactions of the American Mathematical Society},
  68(3):337--404, 1950.

\bibitem{brukut22}
Steven~L. Brunton and J.~Nathan Kutz.
\newblock {\em Data-Driven Science and Engineering: Machine Learning, Dynamical
  Systems, and Control}.
\newblock Cambridge University Press, 2nd edition, 2022.

\bibitem{caponnetto_optimal_2007}
Andrea Caponnetto and Ernesto De~Vito.
\newblock Optimal {Rates} for the {Regularized} {Least}-{Squares} {Algorithm}.
\newblock {\em Foundations of Computational Mathematics}, 7:331--368, 2007.

\bibitem{carmeli_2006}
Claudio Carmeli, Ernesto De~Vito, and Alessandro Toigo.
\newblock Vector valued reproducing kernel hilbert spaces of integrable
  functions and mercer theorem.
\newblock {\em Analysis and Applications}, 04(04):377--408, 2006.

\bibitem{cucker_foundations_2002}
Felipe Cucker and Steve Smale.
\newblock On the mathematical foundations of learning.
\newblock {\em Bulletin of the American Mathematical Society}, 39(1):1--49,
  2002.

\bibitem{devore_sup_2006}
Ronald DeVore, Gerard Kerkyacharian, Dominique Picard, and Vladimir Temlyakov.
\newblock Approximation methods for supervised learning.
\newblock {\em Foundations of Computational Mathematics}, 6(1):3--58, 2006.

\bibitem{devroye_precog_1996}
Luc Devroye, László Györfi, and Gábor Lugosi.
\newblock {\em A Probabilistic Theory of Pattern Recognition}.
\newblock Springer, 1996.

\bibitem{dudley_prob_2002}
R.~M. Dudley.
\newblock {\em Real Analysis and Probability}.
\newblock Cambridge University Press, 2002.

\bibitem{engl_regularization_2000}
Heinz~Werner Engl, Martin Hanke, and A.~Neubauer.
\newblock {\em Regularization of Inverse Problems}.
\newblock Springer, 2000.

\bibitem{giannakis_data_2019}
Dimitrios Giannakis.
\newblock Data-driven spectral decomposition and forecasting of ergodic
  dynamical systems.
\newblock {\em Applied and Computational Harmonic Analysis}, 47(2):338--396,
  2019.

\bibitem{gyorfi_distfree_2002}
László Györfi, Michael Kohler, Adam Krzyżak, and Harro Walk.
\newblock {\em A Distribution-Free Theory of Nonparametric Regression}.
\newblock Springer, 2002.

\bibitem{HoerlKennard1970}
Arthur~E. Hoerl and Robert~W. Kennard.
\newblock Ridge regression: Biased estimation for nonorthogonal problems.
\newblock {\em Technometrics}, 12(1):55--67, 1970.

\bibitem{kostic_learning_2022}
Vladimir Kostic, Pietro Novelli, Andreas Maurer, Carlo Ciliberto, Lorenzo
  Rosasco, and Massimiliano Pontil.
\newblock Learning dynamical systems via koopman operator regression in
  reproducing kernel hilbert spaces.
\newblock In {\em Advances in Neural Information Processing Systems}, 2022.

\bibitem{Kov2024}
Nikola~B. Kovachki, Samuel Lanthaler, and Andrew~M. Stuart.
\newblock Operator learning: Algorithms and analysis.
\newblock {\em Applied and Computational Harmonic Analysis}, 67:101731, 2024.

\bibitem{martinsson_tropp_2020}
Per-Gunnar Martinsson and Joel~A. Tropp.
\newblock Randomized numerical linear algebra: Foundations and algorithms.
\newblock {\em Acta Numerica}, 29:403–572, 2020.

\bibitem{meytwe09}
Sean~P. Meyn and Richard~L. Tweedie.
\newblock {\em Markov Chains and Stochastic Stability}.
\newblock Cambridge University Press, 2nd edition, 2009.

\bibitem{mezic_koopman_2021}
Igor Mezi{\'c}.
\newblock Koopman operator, geometry, and learning of dynamical systems.
\newblock {\em Notices of the American Mathematical Society}, 68(7):1087--1105,
  2021.

\bibitem{micchelli_2005}
Charles~A. Micchelli and Massimiliano Pontil.
\newblock On learning vector-valued functions.
\newblock {\em Neural computation}, 17(1):177--204, 2005.

\bibitem{rudi_less_2015}
Alessandro Rudi, Raffaello Camoriano, and Lorenzo Rosasco.
\newblock Less is more: {Nyström} computational regularization.
\newblock In {\em Advances in Neural Information Processing Systems}, 2015.

\bibitem{rudi_falkon_2017}
Alessandro Rudi, Luigi Carratino, and Lorenzo Rosasco.
\newblock {FALKON}: {An} {Optimal} {Large} {Scale} {Kernel} {Method}.
\newblock In {\em Advances in Neural Information Processing Systems}, 2017.

\bibitem{schwartz_vec_1964}
L.~Schwartz.
\newblock Sous-espaces hilbertiens d'espaces vectoriels topologiques et noyaux
  associ{\'e}s (noyaux reproduisants).
\newblock {\em J. Analyse Math.}, 13:115--256, 1964.

\bibitem{SmaleZhou2004}
Steve Smale and Ding-Xuan Zhou.
\newblock Shannon sampling and function reconstruction from point values.
\newblock {\em Bull. Amer. Math. Soc. (N.S.)}, 41(3):279--305, 2004.

\bibitem{smola_sparse_2000}
Alex~J. Smola and Bernhard Schölkopf.
\newblock Sparse {Greedy} {Matrix} {Approximation} for {Machine} {Learning}.
\newblock In {\em {ICML} 17}, 2000.

\bibitem{TikhonovArsenin1977}
A.~N. Tikhonov and V.~Y. Arsenin.
\newblock {\em Solutions of Ill-posed Problems}.
\newblock Winston \& Sons, Washington, 1977.

\bibitem{vapnik_slt_1998}
Vladimir~N. Vapnik.
\newblock {\em Statistical Learning Theory}.
\newblock Wiley, 1998.

\bibitem{devito_invprob_2005}
Ernesto~De Vito, Lorenzo Rosasco, Andrea Caponnetto, Umberto~De Giovannini, and
  Francesca Odone.
\newblock Learning from examples as an inverse problem.
\newblock {\em Journal of Machine Learning Research}, 6(30):883--904, 2005.

\bibitem{wahba_book_1990}
Grace Wahba.
\newblock {\em Spline Models for Observational Data}.
\newblock SIAM, 1990.

\bibitem{williams_2000}
Christopher Williams and Matthias Seeger.
\newblock Using the {N}ystr\"{o}m method to speed up kernel machines.
\newblock In {\em Advances in Neural Information Processing Systems},
  volume~13, 2000.

\end{thebibliography}


\end{document}